# An Ensemble Deep Convolutional Neural Network Model for Electricity Theft Detection in Smart Grids

Hossein Mohammadi Rouzbahani, Hadis Karimipour, Lei Lei
School of Engineering
University of Guelph
Guelph, Canada
{hmoham15, hkarimi, leil}@uoguelph.ca

*Abstract*—Smart grids extremely rely on Information and Communications Technology (ICT) and smart meters to control and manage numerous parameters of the network. However, using these infrastructures make smart grids more vulnerable to cyber threats especially electricity theft. Electricity Theft Detection (EDT) algorithms are typically used for such purpose since this Non-Technical Loss (NTL) may lead to significant challenges in the power system. In this paper, an Ensemble Deep Convolutional Neural Network (EDCNN) algorithm for ETD in smart grids has been proposed. As the first layer of the model, a random under bagging technique is applied to deal with the imbalance data, and then Deep Convolutional Neural Networks (DCNN) are utilized on each subset. Finally, a voting system is embedded, in the last part. The evaluation results based on the Area Under Curve (AUC), precision, recall, f1-score, and accuracy verify the efficiency of the proposed method compared to the existing method in the literature.

## I. Introduction

Nowadays, power systems are under high pressure to provide a stable and reliable supply of electricity. According to the U.S. Department of Energy, demand for electricity is expected to grow 30% by 2035 as a result of new consumption models (smart plug-in vehicles and smart homes)[1], [2]. Response to this huge demand requires a new generation of power systems, which is also known as the smart grid [3], [4]. Enabled by Advanced Metering Infrastructure (AMI) and communication networks, smart grid computerized the traditional power system into a smart bi-directional system, which increase efficiency, reliability, and security of energy production and distribution. [5], [6]. AMIs are physically unguarded devices that are installed all over the grid to collect data for online monitoring of the system. Although AMIs provide insight into the grid operation and facilitates real-time power system control, it increases the vulnerability of the power system towards intelligently designed cyber/physical attacks., [7], [8].

Electricity theft falls under both physical and cyber-attacks since, on the one hand, an illegal customer can attack the network by bypassing, hacking, or tampering the AMIs physically, and on the other hand, a third party can target the system by a cyber-attack [9]. Cyber-attack can take place by false data injection or hacking utility software by a customer or an insider attacker. The rate of electricity theft, which is categorized as a non-technical loss, is very high all around the world. More than 20% of electricity in India has been consumed by electricity thieves with no payment to the utilities [10]. Also, the cost of this attack in Canada is about 100 million Canadian dollars per year[11].

Load prediction plays a major role in the power system. It can help utilities on the one hand to estimate future electricity demand based on past data and on the other hand help them in decision making, operation scheduling and maintenance planning [12]. Error in the load prediction not only leads to an economic loss but also result in technical aspects [13] since during electricity theft, the power system is facing with an unknown Non-Technical Loss (NTL). For instance, the biggest outage in history that occurred in India was the consequences of electricity theft [14].

Traditional inspection-based Electricity Theft Detection (ETD) techniques cannot be applied to smart grids due to the complexity of the grid and the huge amount of data generated by many users[15]. Therefore, robust and precise methods like machine learning approaches are required to process historical network data and extract patterns to detect any abnormal activities of users [16], [17]. Among different machine learning algorithms, Support Vector Machine (SVM) and Artificial Neural Network (ANN) are showing promising results for anomaly detection in smart grids. Nagi et al. [18] applied SVM for anomaly behavior detection based on the customer's load profile, and they improved the rate by 53%. Later in [19] they introduced a fuzzy inference system, which increased the detection rate up to 72%. Jokar et al. [20] developed an ETD algorithm, which is a combination of clustering and SVM, where clustering was applied as a primary step. ANN was also used for EDT by Glauner et al [21]. Jindal et al. [22] proposed a Decision Tree and SVM-Based method to detect and locate real-time electricity theft in smart grids. Random forest is another method that has been conducted in [23]. Aydin et al. [24] developed a K-Nearest

Neighbor (KNN) for non-technical electricity loss detection in commercial sections. Singh et al. [25] proposed KNN with Principal Component Analysis (PCA) to detect energy theft in AMI. Punmiya et al. [15] presented a Gradient Boosting Theft Detector (GBTD) algorithm with the focus on feature engineering.

To increase the accuracy of ETD by artificial feature extraction, authors in [26] proposed a detection technique based on the Convolutional Neural Networks (CNN). Also, a combination of a long short-term memory (LSTM) architecture and CNN has been develop in [27] by using a synthetic minority over-sampling technique (SMOTE). Although, the accuracy has been improved in two last mentioned studies, but there is still a lot to improve on the detection algorithm performance. There are many vital aspects of a detection algorithm that require more subtlety, excluding dealing with missing values, handling outliers, detecting actual malicious behaviors and carrying highly imbalanced dataset.

The existing ETD techniques suffer from two main problems. First, most of the existing work ignored the fact that the real smart grids data sets are highly imbalanced. A dataset is imbalanced if the instances of some classes are far fewer than other classes. The fundamental principle of classification is finding the boundary between different classes. If some classes are rarely presented, they may not be able to provide enough information to determine the boundary. Therefore, they may be treated as outliers resulting in wrong classifications. To address this issue, this paper proposes a random bagging technique to handle the imbalance dataset by converting the original data into a group of balanced subsets. Secondly, there are several non-malicious factors that can affect consumption patterns including weather conditions and seasonality, install or remove an appliance, change of residents, weekends and holidays, which again are neglected in the literature. These normal activities that lead to a sudden change in the load shape may mistakenly recognize as an abnormal behavior. Since this error may result in the high false-positive rate, feature extraction is vital in the ETD. To overcome this problem, a deep CNN is developed for each dataset.

The Ensemble Deep Convolutional Neural Network (EDCNN) is tested on the real dataset issued by State Grid Corporation of China (SGCC) [28], whilst a new method has been utilized for filling the missing values in this dataset. The consumption pattern not only is dissimilar in different seasons and months also it does not follow the same pattern even on different days of a week. The sensitivity of the load shape to the type of days has been taken into the account. Also, instead of ignoring outliers, a method named Winsorization has been applied for replacing outliers. Since the actual consumption may be zero in some days for many users, it causes a fault in the detection procedure, a negative value has been assigned in lieu of zero to avoid the error.

The key contributions of this research are as follows:

- Proposing a data extrapolation method for data preprocessing to deal with the missing values and outliers.
- Developing an ensemble deep CNN model for ETD in smart grids with high accuracy and low false-positive rate.
- Proposing a random bagging subsampling technique to increase the robustness against imbalanced data by increasing f-measure.

The remainder of this paper is structured as follows. Problem analysis is presented in section II. Section III describes our methodology in this study also including an overview of the dataset. The results are presented in Section IV. Lastly, we conclude the paper in Section V.

II. ENSEMBLE DEEP CONVOLUTIONAL NEURAL NETWORK (EDCNN) ALGORITHM

An anti-social behavior like electricity theft can cause serious problems in smart grids. Although the main purpose of electricity theft is to reduce the electricity bill, this malicious activity puts on a huge burden on the system that may lead to catastrophic damages through a power outage.

Wide deployment of the AMIs in smart grids opens the opportunity for real-time grid monitoring to detect abnormal consumption behavior of the users due to the electricity theft. There are many ETD methods developed in the literature which take advantages of machine learning techniques. However, majority of them ignore the imbalance property of the dataset which affects the election accuracy. Besides that, there are lots of non-malicious events that can be detected as an attack; thereby resulting in a high false-positive rate. In comparison with two recent studies on the same dataset, EDCNN presents more straightforward techniques to deal with (i) missing values considering seasons, months, weekdays and weekends (ii) replacing outliers (iii) distinction between normal and malicious activities and (iv) highly imbalance dataset. To address these issues, in this paper a random bagging technique is proposed to handle the imbalance dataset by converting the original data into a group of balanced subsets. Furthermore, to decrease the false-positive rate, a deep CNN is developed for ETD. Overall, the proposed method includes two main steps: data preprocessing and ETD using EDCNN.

*A. Preprocessing*

More than 75% of the researcher's time in data mining is taken by data processing [29] and one of the major problems is missing values in the dataset. There are many reasons for missing values, including meter failure, power outage, unscheduled maintenance, physical damage to sensors, cyber-attacks, etc.[30] . A new method for filling missing values in this dataset is proposed considering the effect of weekdays, weekends, and different months on consumption pattern. Also, if filling a missing value is impossible by assigning a negative value for these NaN values, the algorithm can understand the difference with real zero. Equation 1 shows our method to recover missing values for a specific day (e.g. Monday):

$$f(C_{M_i}) = \begin{cases} \frac{\sum_{n=1}^{x} C_{M_n}}{n} & 1 \leq n \leq 4, n \in \mathbb{N} \\ C_{M_i} & M_i \notin NaN \\ -1 & n = 0 \end{cases} \quad (1)$$

where $C_{M_i}$ is electricity consumption on the $i^{th}$ Monday in a specific month (there are four or five Mondays in a month). $C_{M_n}$ denotes the amount of consumption on the other Mondays in that specific month (except $i^{th}$ Monday).

The next step is handling enormous consumptions as an outlier. Winsorization is a method to replace an outlier with the nearest acceptable value [31]. We applied the least winsorized square to replace outliers with new estimated consumption.

Finally, the dataset is normalized to decrease the sensitivity of CNN to the dataset diversity. Equation 2 presents MIN-MAX scaling formula that is used in our data preprocessing [32].

$$f(c_i) = \frac{c_i - \text{Min}(C)}{\text{Max}(C) - \text{Min}(C)} \quad (2)$$

### B. Random Under Bagging

After dealing with missing values and outliers, highly imbalanced data is another major challenge since it may lead to misclassification. There are two approaches to dealing with imbalanced data including techniques that concentrate on classifier level or data level [33]. As a machine learning approach, based on the data level, the bagging ensemble of classifiers is a technique to deal with imbalanced data and increasing accuracy. Several bagging approaches can be presumed, including Under Bagging, Over Bagging, Under-Over Bagging, and IIVotes [34].

Random under bagging has been utilized in this study. Each bag contains all samples from the minority class, and the rest of the samples are selected randomly and from the opposite class where the bag is balanced. Even though there is no mandatory instruction for the number of bags, but it is preferred to avoid reducing the total number of the majority class.

### C. Deep Convolutional Neural Network (DCNN)

The problem of imbalanced data has been addressed using random under bagging technique. Analyzing consumption pattern is the fundamental step in ETD since the shape of the load curve for normal and thieve users are different. Also, the algorithm must be able to distinguish between normal pattern changes and abnormal behavior. Neural Networks can robustly extract and learn lots of feature maps using sequential convolutional layers, kernels, and spatial pooling layers. In this work a CNN model for ETD is proposed. All layers in the proposed CNN are fully connected to find the principal features. Linear filters with nonlinear activation functions are applied by convolutional layers that lead to creating feature maps from the dataset. Pooling layers are responsible to avoid overfitting, ignoring redundant features and decreasing the number of parameters. Partial features can be found by CNN in its initial layers and more important feature in farther ones. Lastly, via an activation function, CNNs can recognize classes of identified objects.

The structure of the proposed method is shown in Figure 1, which mainly contains three major components, including under-sampling by random bagging strategy, deep convolutional neural networks, and ensemble via voting classifier.

In summary, at first, the preprocessed dataset has been split up into nine subsets by random bagging method where the sum of normal users is equal to the primary dataset. Each bag of data is a balanced dataset created based on the number of attack samples. Algorithm 1 shows flow of the EDCNN. This method is used based on the environment of the dataset since the only feature for every single day is the total consumption of users. In contrast with other methods, there is no need to generate new samples as it may cause a fault in the detection procedure.

**Algorithm 1**: Procedure of the EDCNN

#Preprocessing
  #Filling Missing Values
    **for**   $C_{M_i}$   $1 \leq M \leq 7$ M: Number of a day in the week
      **if**   $C_{M_i} = NaN$
        **if**   $1 \leq n \leq 4, n \in \mathbb{N} : f(C_{M_i}) = \frac{\sum_{n=1}^{x} C_{M_n}}{n}$
        **else**   $f(C_{M_i}) = -1$
      **else**   $f(C_{M_i}) = C_{M_i}$
  **end loop**
  #Outliers Replacement
    **Using Least-Winsorized-Square**
  #Scaling
    $f(c_i) = \frac{c_i - \text{Min}(C)}{\text{Max}(C) - \text{Min}(C)}$
# Split up the imbalanced dataset to nine balance subsets
    **Using Random Under Bagging**
#Applying DCNN on each subset
    **Split data for Train (70%) and Test (30%)**
    **Reshape data to 3-D**
    **Set number of layers**
    **Set: Number of filters, Kernel size, Activation function**
    **Pooling (Max-Pooling) - Set pool size**
    **Flatting – Activation Function**
#Ensemble Model
    $X = Number\ of\ \{y_i = Attack\}$
    $Y = Number\ of\ \{y_i = Normal\}$
      **for** i in range L   #L= {Number of sets}, L%2≠0
        **if**   $X > Y$   **then**:   $y_i = Attack$
        **else**   $y_i = Normal$
    **end loop**
#END

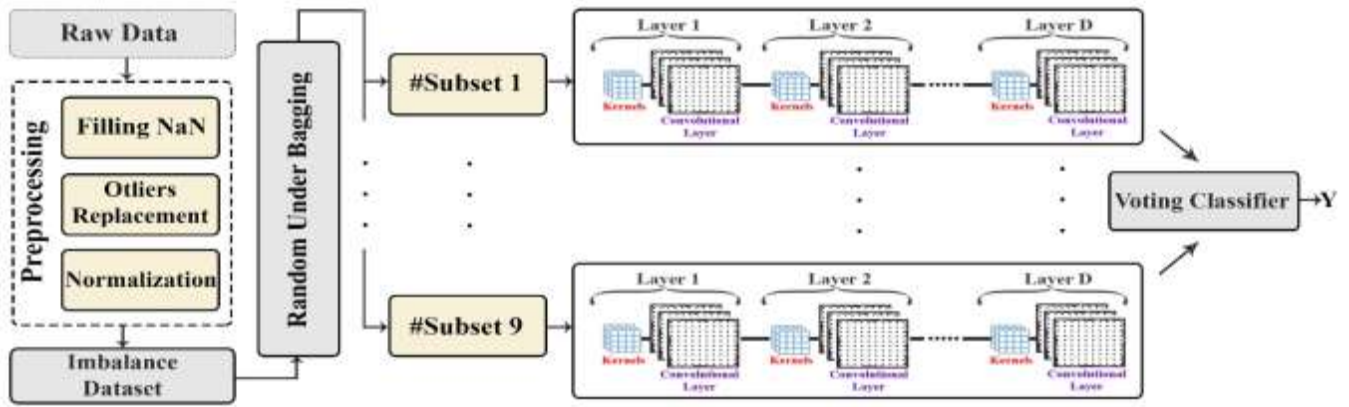

Fig 1. The structure of EDCNN

## III. RESULTS

The proposed algorithm has been tested on a real data set collected by State Grid Corporation of China (SGCC) [28] and The dataset consists of daily consumption of 42372 users over a period of two years (January 1,2014 – December 31, 2015). In the dataset, only 8% of customer are attackers, and the rest are normal users that shows the dataset is highly unbalanced. The test setup uses Python 3.7.4 on a standard system with an Intel Core i7-97580H CPU with 16.0 GB of RAM. The model's design is structured based on TensorFlow.

Performance of the proposed method is analyzed based on the confusion matrix parameters. Table 1 shows a confusion matrix where True Positive (TP), True Negative (TN), False Positive (FP), False Negative (FN) represent the number of customers that are categorized correctly as normal, classified correctly as attacker, categorized falsely as normal and classified falsely as attacker respectively [35].

Table 1. Confusion Matrix

| Actual/Detected | Normal | Attacker |
|---|---|---|
| Normal | TP | FN |
| Attacker | FP | TN |

There are four key performance parameters that can be obtained from the confusion matrix named Precision, Recall, AUC, and F1-score [36]. Precision is the ratio of users that correctly detected as normal per entire customers, which are detected as normal. Recall symbolizes the ratio of correctly detected positive users per the number of all users. AUC specifies the worth of the classifier, and finally, equation 3 demonstrates f1-score mathematically.

$$f1-score = \frac{2 \times Precision \times Racall}{Precision + Racall} \quad (3)$$

Initially, we tried different classifiers on the dataset to make a comparison between their performances for ETD. Table 2 shows the confusion matrix for the EDCNN model. As can be seen from the Table 2, the propose method results is a very low false-positive rate which is highlight important for ETD. For further analysis, the proposed method is compared with other existing work and base classifiers including SVM, Gradient Boosting Model (GBM), Random Forest, and DCNN. Table 3 shows a summary of the results. As can be seen from the result, the proposed method outperforms other techniques in all evaluation metrices. Also, the performance of the EDCNN is higher than the two other algorithms that have been proposed in [26] and [27].

Table 2. Confusion Matrix for the Proposed Model

| Actual/Detected | Normal | Attacker |
|---|---|---|
| Normal | 11536 | 140 |
| Attacker | 109 | 1085 |

Table 3. Results Comparison

| Classifier | AUC | Precision | Recall | F-score |
|---|---|---|---|---|
| SVM | *0.788* | *0.624* | *0.552* | *0.585* |
| GBM | *0.815* | *0.693* | *0.647* | *0.669* |
| Random Forest | 0.892 | 0.911 | 0.751 | 0.823 |
| DCNN | 0.883 | 0.901 | 0.842 | 0.871 |
| Wide-deep CNN [26] | 0.72 | 0.96 | - | - |
| CNN-LSTM [27] | - | 0.90 | 0.91 | 0.89 |
| EDCNN | *0.993* | *0.988* | *0.990* | *0.989* |

The proposed EDCNN can perfectly detect electricity theft in smart grids while the model's AUC and f-score are 0.993 and 0.989, respectively. Also, the reported accuracy for the EDCNN is 0.981. The trend of accuracy and loss, (for train and test stages over 50 epochs) is shown in Fig. 3 and Fig. 4, respectively. As indicated in Figure 3, after 10 epochs the accuracy is fit into an acceptable range. Since the dataset has become completely balanced, accuracy can indicate the correctness of the algorithm. Also, Figure 4 shows there is no overshoot in loss after epoch 25 and the trend of this figure signifies that the model works well without overfitting.

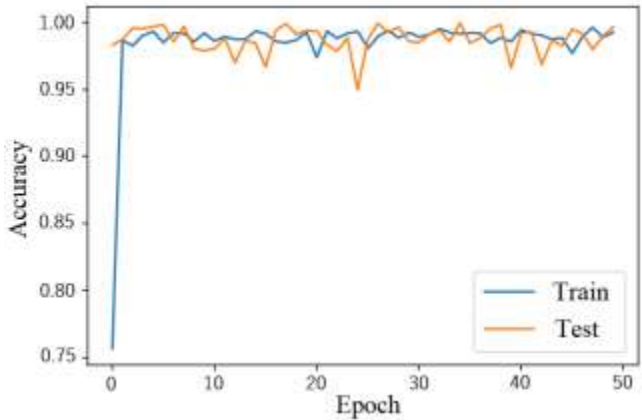

Fig 3. Accuracy of the EDCNN

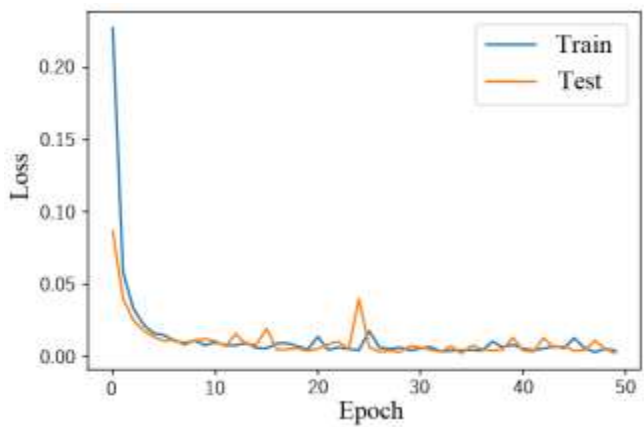

Figure 4. Accuracy of the EDCNN

## IV. Conclusion

Electricity theft declines the performance of smart grids. Insofar in worst case as may lead to system collapse. In this paper an ensemble CNN model for ETD is proposed. To deal with the imbalanced property of the dataset, a random bagging method is applied to create balance subsets. Each subset has been fed into a deep CNN for ETD. Finally, majority voting is used for attack detection. The proposed method is tested on a real dataset containing more than 42000 users over a period 0f 24 months.

The results verify the efficiency of the proposed model for ETD with high accuracy and low false-positive.

For future work, we will study the abnormal behavior of users based on their hourly consumption to detect electricity theft. Also, we aim to develop a model to detect insider attacks on the system, which has not been well addressed.